\documentclass[letterpaper, 10 pt, journal]{IEEEtran}  

\IEEEoverridecommandlockouts                              

\usepackage{graphicx}
\usepackage{mathtools, cuted}
\usepackage{booktabs}
\usepackage{comment}
\usepackage{graphics}
\usepackage[table,xcdraw]{xcolor}
\usepackage{multicol}
\usepackage{algorithm}
\usepackage{lmodern,amsmath,amssymb}
\usepackage[noend]{algpseudocode}
\usepackage{caption}
\usepackage{paralist}
\usepackage{subcaption}
\usepackage{setspace}
\usepackage{wrapfig}
\usepackage[hyphens]{url}

\algnewcommand\algorithmicforeach{\textbf{for each}}
\algdef{S}[FOR]{ForEach}[1]{\algorithmicforeach\ #1\ \algorithmicdo}

\title{\LARGE \bf
Leveraging Domain Knowledge using Machine Learning for Image Compression in Internet-of-Things}

\author{\large Prabuddha Chakraborty, Jonathan Cruz, and Swarup Bhunia
\\  Department of Electrical \& Computer Engineering
\\University of Florida, Gainesville, FL, USA}

\begin{document}

\maketitle
\thispagestyle{empty}
\pagestyle{empty}

\begin{abstract}
The emergent ecosystems of intelligent edge devices in diverse Internet of Things (IoT) applications, from automatic surveillance to precision agriculture, increasingly rely on recording and processing variety of image data. Due to resource constraints, e.g., energy and communication bandwidth requirements, these applications require compressing the recorded images before transmission. For these applications, image compression commonly requires: (1) maintaining features for coarse-grain pattern recognition instead of the high-level details for human perception due to machine-to-machine communications; (2) high compression ratio that leads to improved energy and transmission efficiency; (3) large dynamic range of compression and an easy trade-off between compression factor and quality of reconstruction to accommodate a wide diversity of IoT applications as well as their time-varying energy/performance needs. To address these requirements, we propose, MAGIC, a novel machine learning (ML) guided image compression framework that judiciously sacrifices visual quality to achieve much higher compression when compared to traditional techniques, while maintaining accuracy for coarse-grained vision tasks. The central idea is to capture application-specific domain knowledge and efficiently utilize it in achieving high compression. We demonstrate that the MAGIC framework is configurable across a wide range of compression/quality and is capable of compressing beyond the standard quality factor limits of both JPEG 2000 and WebP. We perform experiments on representative IoT applications using two vision datasets and show up to 42.65x compression at similar accuracy with respect to the source. We highlight low variance in compression rate across images using our technique as compared to JPEG 2000 and WebP.

\end{abstract}


\section{Introduction}

In the Internet of Things (IoT) era, humans have been increasingly removed from the surveillance loop in favor of a connected ecosystem of edge devices performing vision-based tasks \cite{ref:al2018survey}. Automatic analysis is the only viable option given the huge amount of data continuously collected from different IoT edge devices. For example, resource-constrained unmanned aerial vehicles (UAVs) or image sensors can be used as surveillance devices for detecting forest fires~\cite{ref:firedrone} or infrastructure damages after natural disasters~\cite{ref:koch2015review}. In these scenarios, autonomous UAVs or edge devices collect data that may be sent to other edge devices or to the cloud for automated machine learning (ML) based analysis. According to the 2019 Embedded Markets Study \cite{EESurvey}, 43\% of IoT applications incorporating advanced technologies are using embedded vision and 32\% are using machine learning. However, using these IoT devices often requires meeting the tight storage, energy and/or communication bandwidth constraints, while maintaining the effectiveness of surveillance.

\begin{figure*}[]
\centering
\includegraphics[scale=0.24]{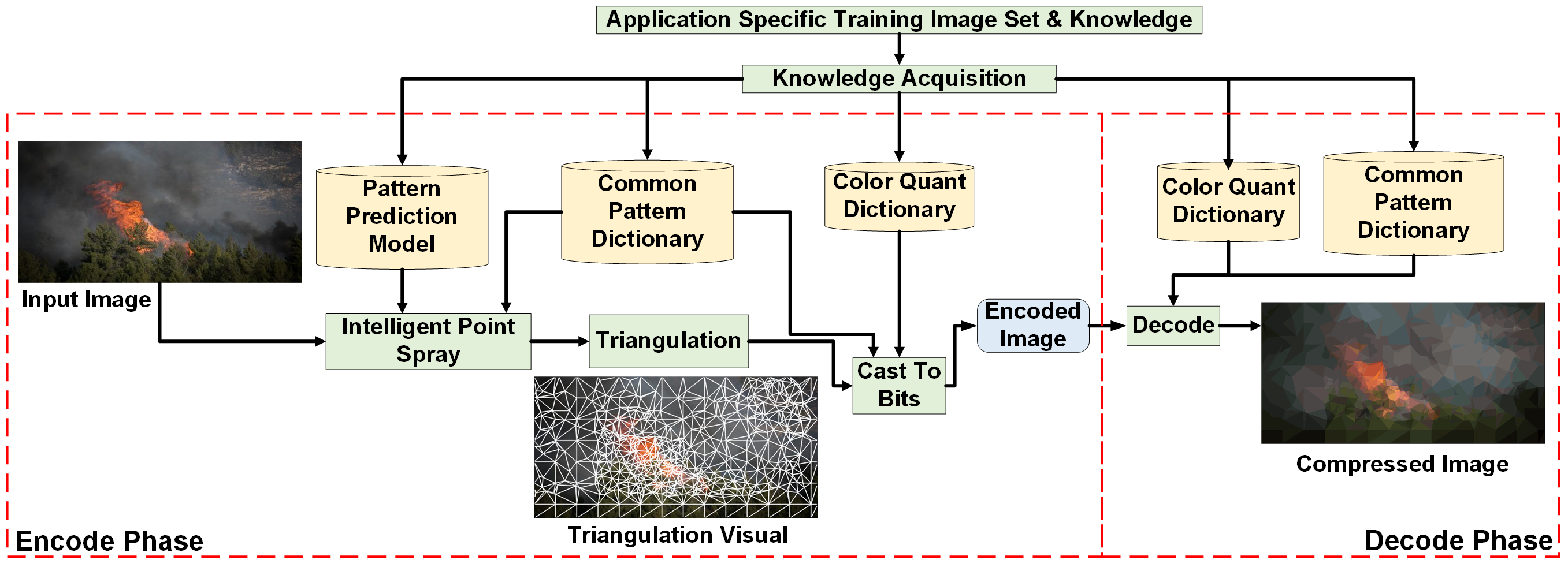}
\caption{Overall flow of MAGIC framework.}
\label{flow}

\end{figure*}

Image compression can address these needs in edge devices that operate in constrained environments and at the same time reduce network traffic \cite{azar2019energy}. Compressed images are easier to store and more energy efficient to transmit long-range. An ideal image compression technique for IoT applications should:
\begin{itemize}
    \item Optimize for machine-to-machine communication and machine-based interpretation in diverse IoT applications - i.e., pattern recognition or feature extraction on the image. Visual perception by human users should be given less importance.
    \item Aim for minimizing the communication bandwidth as IoT is creating 1000X more dense networking requirements \cite{AiIoT1000,iotSpectrumSurvey}, often driven by image/video communication. 
    \item Gear towards minimizing the overall energy and space requirement on resource-constrained edge devices.

\end{itemize}

The standard image compression methods, such as JPEG \cite{ref:jpeg}, JPEG 2000~\cite{ref:jpeg2000},  and WebP~\cite{ref:webp} are tailored to maintain good human-perceivable visual quality and were not designed with IoT applications in mind. Properties of IoT applications which can be leveraged to obtain increased compression are as follows:
\begin{itemize}
    \item The image domain is biased based on the application and on each specific edge image sensor device. The bias can be divided into two categories: (1) color distribution bias, (2) common pattern bias. We define patterns as segment outlines in an image. This information can be learned and utilized. 
    \item Depending on the application, specific entities of the images may hold greater value with respect to the rest of the image. Such applications, therefore, have a region of interest bias which can be learned and utilized. 
    \item Coarse-grained ML tasks prevalent in IoT applications can tolerate extreme levels of compression. 
\end{itemize}

Building on these observations, we propose MAGIC, a \textbf{M}achine le\textbf{A}rning \textbf{G}uided \textbf{I}mage \textbf{C}ompression framework for achieving extreme levels of image compression in IoT systems while maintaining sufficient accuracy for coarse-grained AI tasks. MAGIC consists of three major steps: (1) knowledge acquisition, (2) encoding and (3) decoding. During knowledge acquisition, different application and domain-specific information such as color distribution, common pattern bias and region of interest bias can be extracted in the form of (1) a color quantization dictionary, (2) a common pattern dictionary and (3) a machine learning model which can intelligently represent image segments as a set of common pattern dictionary entries.

During the encoding stage, an image is segmented into non-overlapping triangles using an efficient Delaunay triangulation (DT) method. The ML model, we name pattern prediction model, and the common pattern dictionary from the knowledge acquisition stage are used 
to guide the image segmentation process. 
Finally, the colors are assigned by averaging the pixel colors within each triangle and quantizing them based on the color quantization dictionary, which is constructed by analyzing the color distribution from the domain using k-means.
The decode phase operates similarly by reconstructing the segments using DT and assigning colors from the color quantization dictionary.

We have implemented MAGIC as a completely configurable framework that can be used to compress images from a given dataset. We evaluate MAGIC extensively using two publicly available datasets: fire detection \cite{ref:fireDS} and building crack detection \cite{ref:crackDS} and observe promising performance. For the building crack detection dataset, at a 1.06\% accuracy loss, we obtained 22.09x more compression with respect to the source images. For the fire detection dataset, at a 2.99\% accuracy loss, we obtained 42.65x more compression with respect to the source images. We show up to $\sim$167x more compression than source at a higher accuracy loss ($\sim$13\%).
Furthermore, we analyze the variability in compressed image size and the energy requirements of MAGIC.

The rest of the paper is organized as follows: Section~\ref{sec:related} discusses background of vision in IoT and related works in compression. Section~\ref{sec:motivation} provides motivations for this work and Section~\ref{sec:methodology} introduces the proposed methodology. Section~\ref{sec:results} presents evaluation and comparison of MAGIC with JPEG 2000 and WebP. Section~\ref{sec:discussion} discusses possible extensions and improvements. Section~\ref{sec:conclusion} concludes the paper.

\section{Background \& Related Works}\label{sec:related} 
In this section, we will give a brief introduction to vision tasks in IoT application, and discuss state-of-the-art compression techniques.
\subsection{Computer Vision in IoT Applications}
IoT applications are gaining popularity in several spheres such as industry, home, healthcare, retail, transport and even security \cite{iotFaceDetection}. Many applications in these domains involve capturing images at the edge device and transmitting the image to cloud or other edge devices for analysis. For example:
\begin{itemize}
    \item UAV based fire detection techniques have been proposed which uses optical remote sensing \cite{arialFireDetection}.
    \item Detecting infrastructure damage in a post-disaster scenario using UAV imaging is being investigated in \cite{UAVRoofHole_IoT, UAV_earthquake_IoT}. 
    \item IoT image sensors and computer vision techniques are widely used for flood monitoring, warning and damage mitigation \cite{floodIoT}.
\end{itemize}

Image sensors and intelligent data analysis are two key aspects of surveillance based IoT applications. 
Additionally, security-oriented IoT surveillance applications actively rely on computer vision to detect anomalies\cite{iotFaceDetection}.

\subsection{Need for Image Compression in IoT Vision}
Different IoT applications require sensing image data at the edge and transmitting them over to other edge devices or cloud for analysis. These edge devices operate with strict space, energy, and bandwidth requirements. Compressing images not only has the direct effect of reducing the space and network traffic requirement but also can reduce energy consumption. 

IoT-based communication is expected to reach 50\% of network traffic by 2025 \cite{AiIoT1000}. For example, a typical 4G network is designed to support thousands of devices worth of traffic in a region. However, with the increase in the number of IoT devices being connected to the network, it may become impossible to efficiently serve all devices simultaneously. Therefore, compression at the edge can help reduce network stress. 

 The energy required to transmit data increases with distance \cite{sadler2006data}. 
 For long-range transmission devices such as MaxStream XTend (at 500 mW transmit power), the energy required for one byte of transmission can be higher than 1 million clock cycles worth of computation \cite{sadler2006data}. Hence, even with the cost of additional computation, compression can ultimately lead to less overall energy expenditure.   Due to all these reasons, image compression is a vital step for any IoT vision application.
\begin{figure}[]
\centering
\includegraphics[width=0.9\columnwidth]{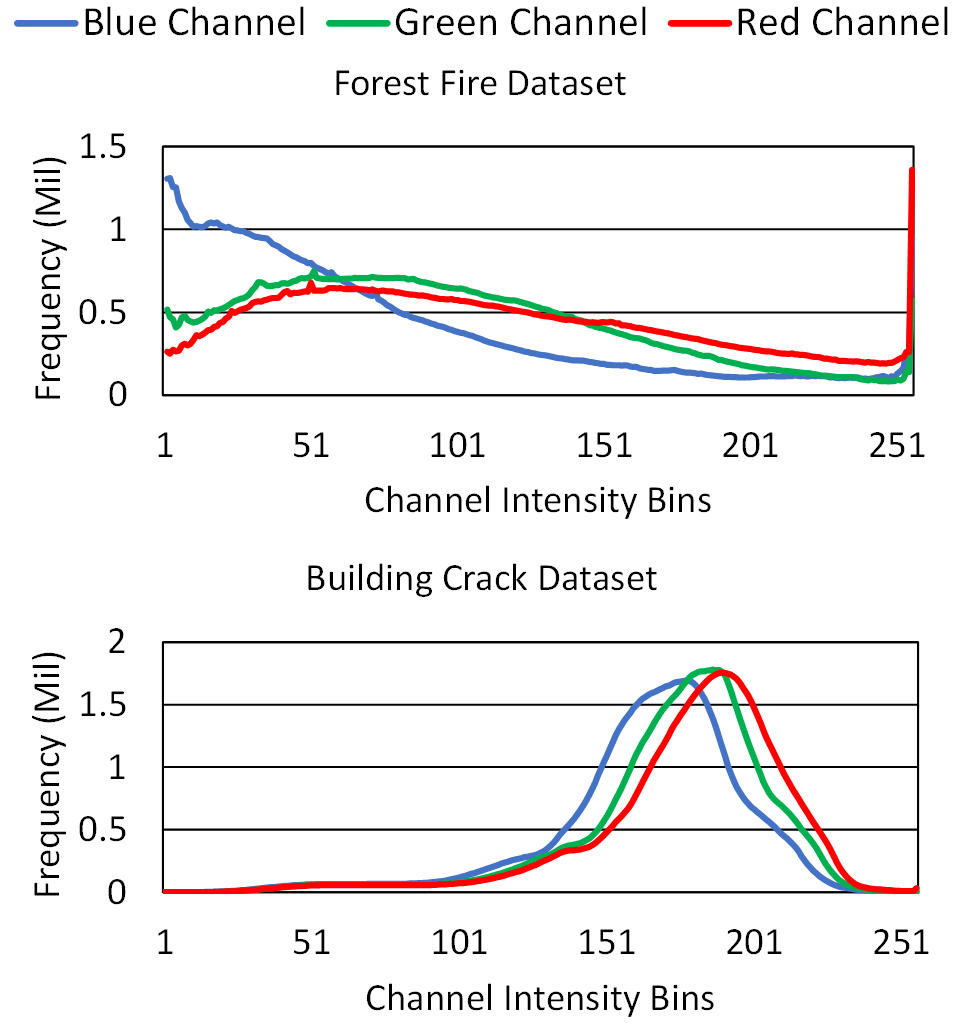}
\caption{Pixel color distribution for forest fire and building crack detection datasets \cite{ref:fireDS,ref:crackDS}. Red, Green and Blue lines represent R,G and B channels respectively.}
\label{color_Dist}
\end{figure}
\begin{figure}[]
\centering
\includegraphics[width=0.9\columnwidth]{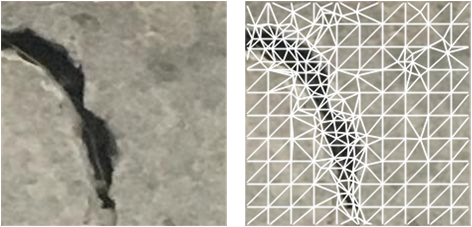}
\caption{DT guided segmentation for a sample building crack detection image \cite{ref:crackDS}.}
\label{CrackSegExample}
\end{figure}
\subsection{State-of-the-art Image Compression Techniques}
\begin{table*}[]
\centering
\caption{Qualitative comparison of MAGIC with different state-of-the-art image compression techniques.}
\label{comparison}
\renewcommand{\arraystretch}{1.3}
\scriptsize\addtolength{\tabcolsep}{-5.5pt}
\begin{tabular}{|c|c|c|c|c|c|c|c|}
\hline
                                               & \textbf{Type}                     & \textbf{Target Application}             & \textbf{\begin{tabular}[c]{@{}c@{}}Domain Knowledge \\ Leveraged\end{tabular}} & \textbf{\begin{tabular}[c]{@{}c@{}}ROI \\ Support\end{tabular}} & \textbf{\begin{tabular}[c]{@{}c@{}}Encoder Space\\ Complexity\end{tabular}} & \textbf{\begin{tabular}[c]{@{}c@{}}Encoder Time \& \\ Energy Requirement\end{tabular}} & \textbf{\begin{tabular}[c]{@{}c@{}}Compression \\ Range\end{tabular}} \\ \hline
\textbf{JPEG 2000\cite{ref:jpeg2000}}                                & Wavelet                           & Human Vision                            & No                                                                             & Yes                                                             & Low                                                                         & Low                                                                                    & Medium                                                                \\ \hline
\textbf{WebP\cite{ref:webp}}                                  & Fequency + Spatial                & Human Vision                            & No                                                                             & No                                                              & Low                                                                         & Low                                                                                    & Medium                                                                \\ \hline
\textbf{Deepn-JPEG\cite{ref:deepnjpeg}}                            & Frequency                         & Complex ML Task                         & Limited                                                                             & No                                                              & Low                                                                         & Low                                                                                    & Medium                                                                \\ \hline
\cellcolor[HTML]{FFFFFF}\textbf{Weber et. al.\cite{ref:weber2019lossy}} & \cellcolor[HTML]{FFFFFF}Frequency & \cellcolor[HTML]{FFFFFF}Complex ML Task & Limited                                                                            & No                                                              & Medium                                                                      & High                                                                                   & Medium                                                                \\ \hline
\cellcolor[HTML]{FFFFFF}\textbf{Marwood et al.\cite{ref:marwood2018representing}} & \cellcolor[HTML]{FFFFFF}Spatial   & \cellcolor[HTML]{FFFFFF}Human Vision    & No                                                                             & No                                                              & Low                                                                         & High                                                                                   & High                                                                  \\ \hline
\textbf{MAGIC}                                 & Spatial                           & Coarse ML Task                          & Yes                                                                            & Yes                                                             & Low                                                                         & Medium                                                                                 &Extreme                                                                  \\ \hline
\end{tabular}
\end{table*}

Several image compression techniques proposed over the years can be divided primarily into two categories: (1) Lossless compression techniques and (2) Lossy compression techniques. Lossless compression techniques such as Arithmetic Coding (\cite{arithmeticCoding_1,arithmeticCoding_2}) and Huffman Coding (\cite{huffman}) aim to completely preserve the content under compression, but generally at the cost of significant mathematical computations \cite{imageCompSurvey}. 
For coarse-grained ML tasks, such quality is not needed, therefore, lossy compression techniques are preferred. As the name suggests, lossy compression allows for variable data loss to achieve higher rates of compression. The data lost is generally not perceivable by humans. Some lossy compression techniques include, JPEG \cite{ref:jpeg}, JPEG 2000 \cite{ref:jpeg2000}, and WebP \cite{ref:webp} which perform quantization in the frequency domain using techniques such as discrete wavelet transform and discrete cosine transform. 
Another class of lossy compression performs quantization in the spatial domain, such as triangulation-based image compression most recently proposed in \cite{ref:marwood2018representing}. This compression technique relies on the DT of a set of points in the image matrix to construct the image out of non-overlapping triangles during both encoding and decoding. In this way, triangulation allows for sending minimal amounts of information at the cost of slightly more encoding/decoding time. While we use DT, our compression algorithm and compression goals are vastly different than \cite{ref:marwood2018representing}.

Machine learning has been used to further improve compression \cite{ref:rippel2017real, ref:toderici2017full, ref:Li_2018_CVPR, ref:balle2018variational}. In all these works, the goal is to maximize visual quality metrics (PSNR, MS-SSIM, and remove artifacts) all while minimizing bits per pixel (BPP). However, complex, large neural networks, are not ideal for use in edge devices. More recently, people have been targeting image compression optimized for ML accuracy over human perceived quality. Liu et al. propose DeepN-JPEG \cite{ref:deepnjpeg} which modifies JPEG's quantization table for deep neural network (DNN) accuracy over human visual quality. DeepN-JPEG is targeted for generalized AI-models and can achieve only 3.5x compression compared to source images. 
However, our approach can achieve up to 42.65x more compression than the source. 
Similarly, in \cite{ref:medical}, Liu et al. modify JPEG 2000 to extract frequencies relevant for neural network (NN) based segmentation of 3D medical images.
Weber et al. develop a recurrent neural network (RNN) based compression with the aim of maximizing the accuracy of generalized classifiers and investigate the accuracies of several classifiers for images compressed for human perception versus machine perception \cite{ref:weber2019lossy}.

In Table~\ref{comparison}, we qualitatively compare MAGIC with different state-of-the-art relevant image compression techniques. MAGIC distinguishes itself as the only image compression technique to be targeted for coarse ML vision tasks in IoT applications. The compression range of MAGIC is higher than other techniques because it is designed to leverage domain knowledge. 

\section{Motivation}
\label{sec:motivation}
Most IoT applications designed to perform a particular automated vision task will have some bias in the images being captured and analyzed. The amount of bias will depend on the application and the sensory edge device in use. For a given application the images will have (1) a pixel color distribution bias depending on the environment where the image occurs and (2) a pattern bias due to prevalence of certain common objects in the images. Apart from the image set bias, the IoT application may have its own bias for certain objects and features which are relevant for the ML analysis task. 

\subsection{Color Distribution Bias}
Image color bias will exist to an extent in any IoT domain-specific application. Apart from the application level color distribution bias, there may be bias attributed to the physical location of the device. Such location bias can be more easily observed for stationary devices. Harnessing the bias for each device separately may be beneficial but in this paper, we limit our study to the application level image color distribution bias. We plot the pixel color distributions for the forest fire dataset \cite{ref:fireDS} and the building crack dataset \cite{ref:crackDS} as shown in Fig.~\ref{color_Dist}. We can clearly observe that certain regions of Red, Green and Blue spectrum are more represented than others. This bias will appear more prominent and severe if we consider the joint Red-Green-Blue distribution. If we could take advantage of this bias by limiting the color space tuned for the specific application then we may be able to compress more.

\subsection{Common Pattern Bias}

\begin{figure}[]
\centering
\includegraphics[width=0.9\columnwidth]{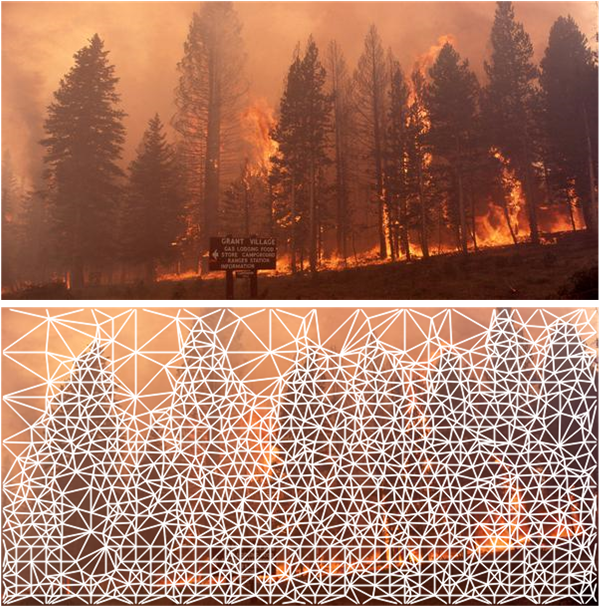}
\caption{DT guided segmentation for a sample forest fire detection image \cite{ref:fireDS}}
\label{fire_seg_example}
\end{figure}
The images captured and analyzed by task-specific IoT applications will have pattern (image segment outlines) bias because of the nature of the objects that are present in the images. For a building crack detection application, the images will consist of cracked and uncracked surfaces (Fig.~\ref{CrackSegExample}) and for a forest fire surveillance application, the images will consist of trees and occasional fires (Fig.~\ref{fire_seg_example}). Just like color distribution bias, common pattern bias will also exist both in the application level and in the device location level. If we could capture and store these domain-specific repeating patterns in a dictionary, for example, then we could potentially save space by storing dictionary entry indices instead of concrete data. 

\subsection{Region of Interest Bias}
Certain objects/regions in the image may hold more importance depending on the IoT task. If the image can be compressed based on the application-specific requirement then we will be able to save important regions at higher quality while sacrificing other regions. For example, let us assume that we have an IoT application which is designed to detect green cars among green and blue cars. Only by using common pattern bias knowledge, we cannot distinguish between green and blue cars. Both cars will have the same level of quality. But with the extra region of interest bias knowledge, we can save space by only learning to only represent the green cars with high quality.

\section{Methodology}\label{sec:methodology} 
\begin{algorithm}[!t]

\scriptsize\addtolength{\tabcolsep}{-1.3pt}
\caption{Knowledge Acquisition}\label{alg:calibration}
\begin{spacing}{1.5}
\begin{algorithmic}[1]
\Procedure{learn}{$bDim, iterLimit, pw, imgList, grid, th, cb$}
\State Initialize $colorFreq = \varnothing, trainX = \varnothing, trainY = \varnothing$
\State $patDict = generatePatternDict(imgList, bDim)$
\ForEach {$img \in imgList $}
    \State $pointArr = \varnothing $
    \State $pointArr = gridSpray(pointArr, grid, img.rows, img.cols)$
    \State $edgePoints = cannyEdgeDetection(img)$
    \State $pointArr.append(edgePoints)$
    \State $iter = 0$
    \While{$iter < iterLimit$}
        \State $pointArr = split(pointArr,img, th)$
        \State $iter = iter + 1$
    \EndWhile
    \State $prunePoint(pointArr, pw)$ \Comment{In every (pw X pw) window,\\\hspace{150pt} keep maximum 1 point}
    \State $triangleList = delaunay\_triangulation (pointArr)$
    \ForEach {$ t \in triangleList $}
        \State $avg\_color = findAvgColor(t, img)$
        \If{$ avg\_color $ in $ colorFreq$}
            \State $colorFreq[avgColor] = colorFreq[avgColor] + 1$
        \Else
            \State $colorFreq[avgColor] = 1$
        \EndIf
    \EndFor
    \State $blockList =  tiling(img, bDim$)
    \State $j = 0$
    \While{$j < length(blockList)$}
        \State $dictInd = assignDictInd(blockList, j, pointArr, patDict)$
        \State $trainX.append(blockList[j])$
        \State $trainY.append(dictInd)$
        \State $j = j + 1$
    \EndWhile
\EndFor
\State $colorDict =weighted\_kmean(colorFreq, k = 2^{cb})$
\State $model = train\_Point\_Prediction\_Model(trainX, trainY)$

\State \textbf{return}  $colorDict, model, patDict$

\EndProcedure
\end{algorithmic}
\end{spacing}

\end{algorithm}

In this section, we present our learning guided compression technique (MAGIC) targeted for coarse-grained ML vision tasks in intelligent IoT ecosystems. Fig.~\ref{flow} illustrates the overall flow. Just like any other compression technique, there is a procedure for encoding the image and a procedure for decoding the image. Additionally, to take advantage of the bias present in the application domain,
we propose a knowledge acquisition procedure. In this paper, we focus on the first aspect of domain knowledge learning, namely, color distribution bias. The other two areas of domain knowledge (pattern bias and ROI bias) are not strictly learned. The common pattern dictionary (for segmentation bias) is statically generated and the pattern prediction model (for ROI bias) is trained based on automated supervision. However, the algorithms are implemented such that future inclusion of human supervision and learning in the other two domain knowledge areas can be easily performed. We will now describe the three major steps of MAGIC in greater detail.

\subsection{Knowledge Acquisition}

Before compression is carried out, the knowledge acquisition procedure is used to analyze a set of sample images from the given use-case and learn common features that can be reused during compression. This learning stage allows for more efficient image compression. To capture the application-specific domain knowledge we use the following constructs and techniques.

\subsubsection{Color Quantization Dictionary}
 We construct a dictionary of most frequently occurring colors for a specific application. Colors are now represented as entries in the dictionary instead of the standard 24-bit RGB value. The number of entries in the dictionary can be controlled by the user. To construct the color dictionary, we first extract the color distribution from a set of domain-specific sample images and then apply unsupervised machine learning (k-means) to extract the colors which are strong representatives of the entire color space. The color quantization dictionary will be used during the encoding and decoding phase for representing the image. Algo. \ref{alg:calibration} describes in details how the color quantization dictionary is constructed. 
 
 \subsubsection{Common Pattern Dictionary}
 Compressing an image with MAGIC involves 
 segmenting an image into representative triangles using Delaunay triangulation (DT). The triangle segments are determined from the points sprayed on the 2D image plane. Hence, patterns in an image segment can be represented as a set of points in a 2D plane. The forest fire images in Fig.~\ref{fire_seg_example} illustrate this process. 
 The common pattern dictionary is a data structure for saving the regularly occurring spray point patterns that occur in an image segment. 
 The patterns are indexed in the dictionary such that a higher index is associated with more complex details. The pattern dictionary can be statically generated to increase compression robustness across different image domains or learned during the knowledge acquisition phase to be in more tune with the application domain.

\begin{algorithm}[!t]

\scriptsize\addtolength{\tabcolsep}{-1.3pt}
\caption{Triangle Split}\label{alg:splitTriangle}
\begin{spacing}{1.5}
\begin{algorithmic}[1]
\Procedure{split}{$pointArr,img, th$}
\State $triangleList = delaunay\_triangulation (pointArr)$
\ForEach {$ t \in triangleList $}
    \State $stdDevColor = calculate\_Color\_Std\_Dev(img,t)$
    \If{$ stdDevColor > th$}
        \State $pointArr.append(barycenter(t))$
    \EndIf
\EndFor
\State \textbf{return}  $pointArr$
\EndProcedure
\end{algorithmic}
\end{spacing}

\end{algorithm}

\subsubsection{Machine Learning Model for Pattern Prediction}
We train a machine learning model that learns to represent the segments of an image as a set of patterns from the Common Pattern Dictionary. Similar to other compression technique, we operate on `blocks' of an image and must partition the image. Each block needs to be assigned a point spray pattern entry from the common pattern dictionary during encoding. The assignment can be based on how much texture details the image block has or the importance of the image block for a given application. MAGIC employs the trained ML model (pattern prediction model) for assigning an image block to an entry from the common pattern dictionary. 

Iterative heuristic driven DT segmentation methods have time complexity $\mathcal{O}(IM\log M)$, where $I$ is the number of iterations and $M$ is the maximum number of points used for computing DT. Our pattern prediction model can provide the points in $\mathcal{O}(1)$ followed by a single DT of complexity $\mathcal{O}(M\log M)$. Therefore, the pattern prediction model has two benefits: (1) The ML guided assignment of an image block to a specific pattern dictionary entry is faster than determining the segmentation pattern of the image block using iterative heuristic means and (2) the ML model can be trained to retain more details for specific image blocks which may be important for the specific visual task.

\subsubsection{Knowledge Acquisition Algorithm}

Before communication can start between a sender entity and a receiver entity, we must construct the above three components during the knowledge acquisition phase. The pattern prediction model (1) must reside on the sender (encoder) side. The common pattern dictionary (2) and color quantization dictionary (3) should reside on both sender and receiver sides.

Algo. \ref{alg:calibration} defines the knowledge acquisition process which can be used to construct these components. We collect a set of sample images (learning dataset) that can approximately represent the nature of images that are to be communicated. In line 3, the common pattern dictionary is generated. For this iteration of MAGIC, the generation is such that entry indexed $i$ has exactly $i$ points sprayed randomly in a ($bDim$ x $bDim$) block. For each image, we construct the $pointArr$ (set of points on the 2D image plane) which determines the segmentation. The $pointArr$ is initially populated with grid points sprayed uniformly based on the parameter $grid$ (line 6 using Algo.~\ref{alg:gridSpray}) and edge points determined by an edge detection algorithm (line 7). In our case, we use canny edge detection. We add more points to the $pointArr$ by repeatedly splitting triangles with standard deviation of pixel intensity greater than $th$ (lines 10-12 using Algo.~\ref{alg:splitTriangle}). This process is done to capture more information, but we note that this may in some cases result in unnecessary details and ultimately less compression. Therefore, we keep at most 1 point in the $pointArr$ for every ($pw$ x $pw$) non-overlapping window (line 13). We then perform DT to obtain the triangle list (line 15). For each triangle in the triangle list, we obtain the average color and update the $colorFreq$. The $colorFreq$ holds the frequency of each triangle color encountered across all the images (lines 16-21). $cb$ (number of bits for representing colors) is a user input to control the size of the color quantization dictionary. We divide the image into blocks of dimension ($bDim$ x  $bDim$) and compute the common pattern dictionary ($patDict$) entry index which best corresponds to the point spray pattern of each block (line 25). The $dictInd$ and the RGB block ($blockList[j]$) act as the label and input data (respectively) for training our point prediction model (lines 26-27). We cluster the entries (weighted by their frequency) in the $colorFreq$ using k-means algorithm \cite{scikit-learn}. The number of clusters is $2^{cb}$. The cluster representatives are assigned an index and collectively form the color quantization dictionary ($colorDict$). In this way, we employ unsupervised machine learning to leverage domain-specific color distribution information. 
The model training process depends on the ML model architecture selected for the domain-specific point prediction task. After the knowledge acquisition phase completes the application is ready to encode (compress) and decode images.

\begin{algorithm}[!t]

\scriptsize\addtolength{\tabcolsep}{-1.3pt}
\caption{Grid Spray Points}\label{alg:gridSpray}
\begin{spacing}{1.5}
\begin{algorithmic}[1]
\Procedure{gridSpray}{$pointArr, grid, rows, cols$}

\State $i = 0$
\While{$i < rows$}
    \State $j = 0$
    \While{$j < cols$}
        \State $pointArr.append((i,j))$
        \State $j = j + grid$
    \EndWhile
    \State $i = i + grid$
\EndWhile
\State \textbf{return}  $pointArr$   
\EndProcedure
\end{algorithmic}
\end{spacing}

\end{algorithm}
\subsection{Encoding Procedure}
\begin{algorithm}[!t]

\scriptsize\addtolength{\tabcolsep}{-1.3pt}
\caption{Image Encoding}\label{alg:encode_main}
\begin{spacing}{1.5}
\begin{algorithmic}[1]
\Procedure{encode}{$bDim, d, img, model, colorDict, patDict, grid$}
\State $block\_list =  tiling(img, bDim$)
\State Initialize $pointArr = \varnothing, labelsArr = \varnothing, bIndex = 0 $
\ForEach {$block \in blockList $}
    \State $label = (predict(block, model, bDim)) / d$
    \State $labelsArr.append(label)$
    \State $points = patDict[label]$
    \ForEach {$p(r,c) \in points $}
        \State $p.c = p.c + (bIndex \% bDim) * bDim $
        \State $p.r = p.r + (bIndex / bDim) * bDim $
    \EndFor
    \State $pointArr.append(points)$
    \State $bIndex = bIndex + 1$
\EndFor
\State $pointArr = gridSpray(pointArr, grid, img.rows, img.cols)$
\State $triangleList =  delaunay\_triangulation (pointArr)$
\State $colorList = \varnothing $
\ForEach {$ t \in triangleList $}
    \State $avgColor = findAvgColor(t, img)$
    \State $quantColor = findClosestMatch(avgColor, colorDict)$
    \State $colorList.append(quantColor)$
\EndFor
\State $encImg = cast\_to\_bits(img.rows, img.cols, grid, bDim,$ \\ \hspace{150pt} $labelsArr, colorList)$
\State \textbf{return}  $encImg$

\EndProcedure
\end{algorithmic}
\end{spacing}

\end{algorithm}

Algo. \ref{alg:encode_main} defines the image encoding process at the sender side. For the given image, we divide it into blocks based on the dimension specified by $bDim$ (line 2). For each block, we predict the pattern dictionary entry to use with the help of the point prediction model (line 5). The label predicted by the ML model is divided by the input $d$, a tunable parameter that allows for dynamic image quality. Higher values of $d$ are associated with higher compression rates. The predicted labels for each block are appended to the $labelsArr$ (line 6). For a label predicted for a specific block, we fetch the associated point spray pattern from the common pattern dictionary ($patDict$) and append the points to the $pointArr$ after computing their absolute position with respect to the image (lines 8-11). $pointArr$ is next populated with grid points sprayed uniformly based on the parameter $grid$ (lines 13 using Algo.~\ref{alg:gridSpray}). We perform DT to obtain the $triangleList$ in line 14. For each triangle in the $triangleList$ we compute the average color ($avgColor$) and find its closet match ($quantColor$) from the color quantization dictionary ($colorDict$). The $quantColor$ is appended to the $colorList$. The final encoded image consists of the following converted and packed as bits:

\begin{itemize}
    \item $img.rows$: The number of pixel rows in the image (16~bits).
    \item $img.col$: Number of pixel columns in the image (16~bits).
    \item $grid$: Number of pixels to skip between 2 grid points sprayed (16~bits).
    \item $bDim$: Dimension of the image block to use (16~bits).
    \item $labelsArr$: $\log_{2}$($patDict$ size) bits for each entry. 
    \item $colorDict$: $\log_{2}$($colorDict$ size) bits for each entry.
\end{itemize}
The encoded image ($endImg$) is returned.

\subsection{Decoding Procedure}
\begin{algorithm}[!t]

\scriptsize\addtolength{\tabcolsep}{-1.3pt}
\caption{Image Decoding}\label{alg:decode_main}
\begin{spacing}{1.5}
\begin{algorithmic}[1]
\Procedure{decode}{$encImg, colorDict, patDict$}
\State $rows, cols, grid, bDim, labelsArr, colorList $=$unpack(encImg)$
\State Initialize $bIndex = 0, pointArr = \varnothing$
\ForEach {$label \in labelsArr $}
    \State $points = patDict[label]$
    \ForEach {$p(r,c) \in points $}
        \State $p.c = p.c + (bIndex \% bDim) * bDim $
        \State $p.r = p.r + (bIndex / bDim) * bDim $
    \EndFor
    \State $pointArr.append(points)$
    \State $bIndex = bIndex + 1$
\EndFor
\State $pointArr = gridSpray(pointArr, grid, rows, cols)$
\State $triangleList = delaunay\_triangulation (pointArr)$
\State $i = 0$
\State $recImg = Array\_of\_Zeros\_of\_Dimension (rows,cols)$
\While {$ i < size(triangleList)$}
    \State $trueColor = colorDict[colorList[i]]$
    \State $drawTriangle(triangleList[i], trueColor, recImg)$
\EndWhile
\State \textbf{return}  $recImg$

\EndProcedure
\end{algorithmic}
\end{spacing}

\end{algorithm}

Algo. \ref{alg:decode_main} defines the image decoding process at the receiver side. Based on the encoding format, $rows$, $cols$, $grid$, $bDim$, $labelArr$ and $colorList$ are extracted from the encoded image ($encImg$) in line 2. For each label in the $labelArr$, we fetch the associated point spray pattern from the pattern dictionary and append the points to the $pointArr$ after computing their absolute position with respect to the image and the block index ($bIndex$) (lines 6-8). The $pointArr$ is next populated with grid points sprayed uniformly based on the parameter $grid$ (line 11 using Algo.~\ref{alg:gridSpray}). We perform DT to obtain the $triangleList$ in line 12. We initialize a blank image with the obtained dimensions in line 14. For each triangle in the $triangleList$, we obtain the RGB color ($trueColor$) from the color quantization dictionary using the corresponding entry from the $colorList$ (line 16). We color the pixels in $recImg$ for the given triangle using $trueColor$ (line 17). The final decoded/recovered image ($recImg$) is returned from this method.

\section{Results}\label{sec:results}

\begin{figure}[]
\centering

\includegraphics[width=0.9\columnwidth]{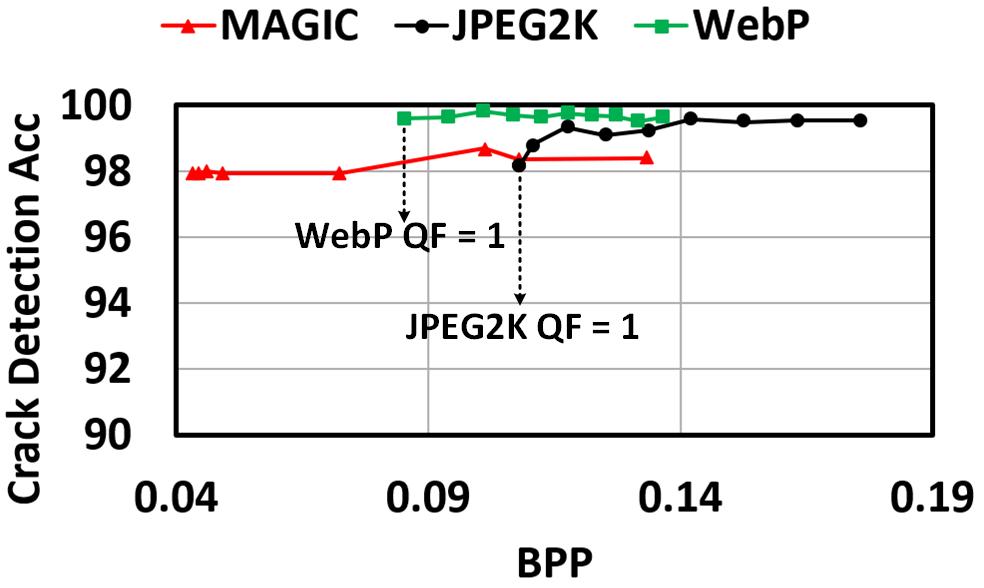}
\caption{Comparison of building crack detection accuracy vs BPP of JPEG 2000, WebP, and MAGIC~(proposed).}
\label{CrackAcc}
\end{figure}
\begin{figure}[]
\centering
\includegraphics[width=0.9\columnwidth]{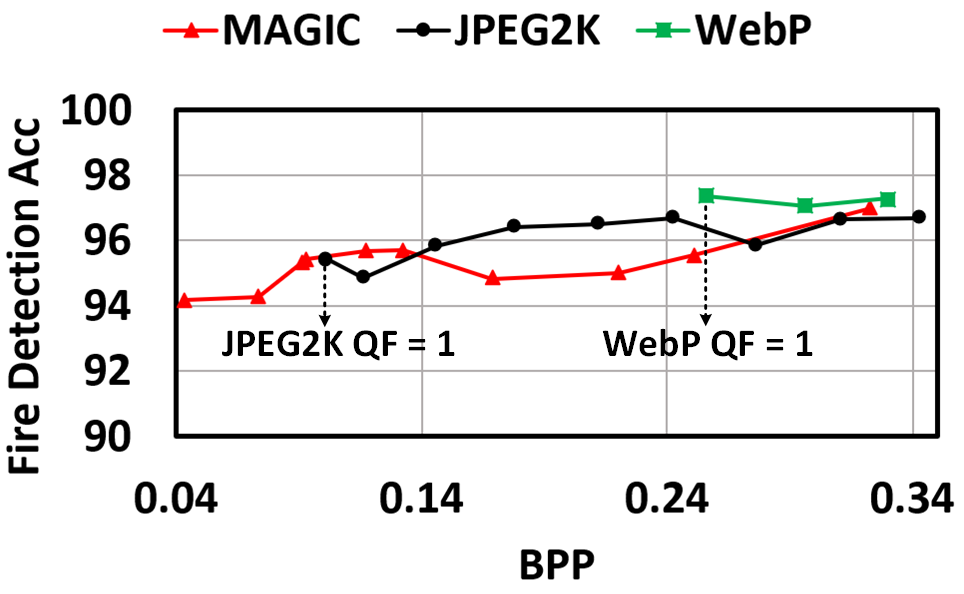}
\caption{Comparison of fire detection accuracy vs BPP of JPEG 2000, WebP, and MAGIC~(proposed).}
\label{fireAcc}

\end{figure}
MAGIC compression is designed to excel in autonomous task-specific IoT applications where the analysis of the images is done by machine learning models. To quantitatively analyze the effectiveness of MAGIC for IoT applications we pick two use-cases:
\begin{enumerate}
    \item \textbf{Forest fire surveillance}~\cite{ref:fireDS}.
    \item \textbf{Infrastructure analysis}~\cite{ref:crackDS}.

\end{enumerate}
 
In the next few subsections,  we describe the experimental setup and compare the accuracy of MAGIC compressed images to JPEG 2000 and WebP under different quality factor (QF) settings.    
We use ImageMagick's convert command for JPEG 2000 and WebP compressions which have quality factor from 1 to 100, with 1 resulting in the highest compression \cite{imageMagick}.  We explore the effect the MAGIC input parameters $pw$, $d$, and $cb$ have on the rate of compression and accuracy. Finally, we introduce a computation, transmission energy cutoff for analyzing the energy efficiency of MAGIC.
\subsection{Experimental Setup}

\begin{figure}[]
\centering
\includegraphics[width=0.95\columnwidth]{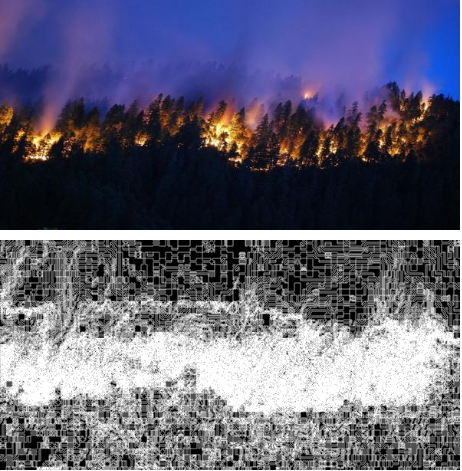}
\caption{Entropy feature of a sample forest fire dataset \cite{ref:fireDS} image visualized by scaling the values between 0 and 255. Brighter pixels have higher entropy.}
\label{EntropyFeatureExample}
\end{figure}
\begin{figure}[]
\centering
\includegraphics[width=0.7\columnwidth]{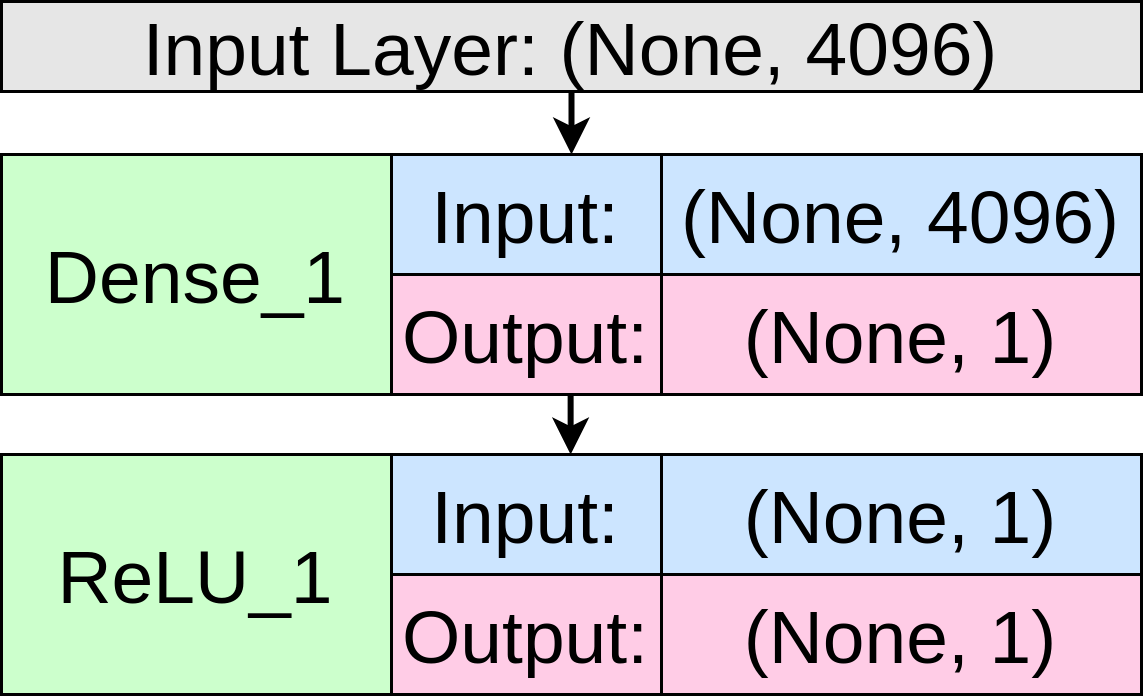}
\caption{Lightweight neural network architecture being used for spray point prediction.}
\label{magicEntropyUltraSmallNet}

\end{figure}

The neural network architecture for the domain-specific ML models is shown in Fig.~\ref{magicEntropyUltraSmallNet}. We obtain separate model weights by training on each dataset and knowledge acquisition parameters (controlling the level of compression) using Keras \cite{chollet2015keras}. The input to the neural network is the flattened per-pixel local entropy features of the 64x64 image blocks. The entropy of a pixel is defined as the number of bits required to represent the local grey-scale distribution of a defined neighbourhood \cite{scikit-image}. A higher entropy value is correlated to higher diversity and higher information density. We use a neighbourhood of 5 pixels to train our models. In Fig.~\ref{EntropyFeatureExample}, we see the visual representation of the entropy feature of a sample image. The output of the neural network domain-specific point prediction model is used to compute the entry in the common pattern dictionary that is to be assigned for the input image block. 

For both building crack detection and forest fire detection task, we use a statically generated point spray pattern dictionary containing 4096 entries such that entry $i$ has exactly $i$ points sprayed randomly in a 64x64 block. Hence using an entry with a high value of $i$ is equivalent to capturing more information in the image block. 
\begin{figure*}[]
\centering
\includegraphics[scale=0.265]{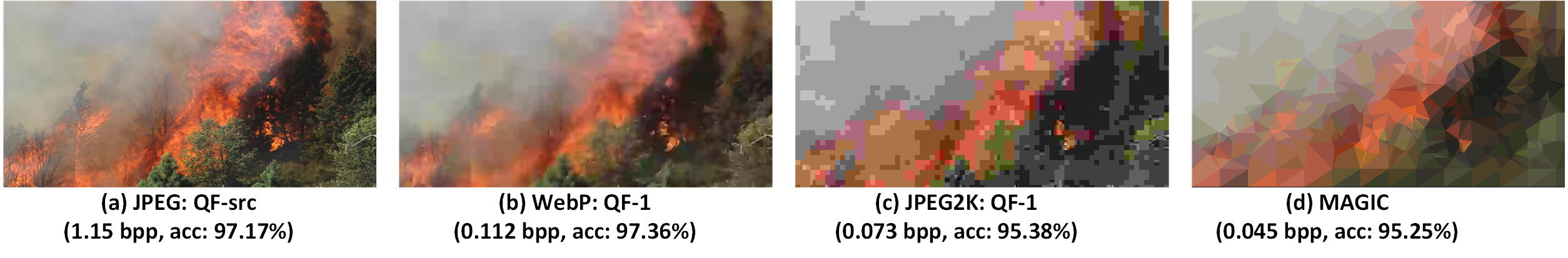}
\caption{Comparison of source, WebP, JPEG 2000, and MAGIC compressed fire detection images.}
\label{fireDS}
\vspace{-0.1in}
\end{figure*}
\subsection{Evaluation up to Lowest Quality Factor }

\begin{table}[]
\centering
\caption{Classification Results for fire dataset \cite{ref:fireDS} and building crack dataset \cite{ref:crackDS} at low and ultra low BPP. }
\label{PushMore}
\renewcommand{\arraystretch}{1.5}
\scriptsize\addtolength{\tabcolsep}{-0.5pt}
\begin{tabular}{ccccccc}
\multicolumn{7}{c}{Ultra Low BPP ($\sim$0.01 BPP)}                                                                                                                                                                     \\ \hline
\multicolumn{1}{|c|}{}             & \multicolumn{3}{c|}{\textbf{Fire DS}}                                                   & \multicolumn{3}{c|}{\textbf{Crack DS}}                                                  \\ \hline
\multicolumn{1}{|c|}{\textbf{BPP}} & \multicolumn{1}{c|}{0.0111} & \multicolumn{1}{c|}{0.0121} & \multicolumn{1}{c|}{0.0135} & \multicolumn{1}{c|}{0.0137} & \multicolumn{1}{c|}{0.0140} & \multicolumn{1}{c|}{0.0211} \\ \hline
\multicolumn{1}{|c|}{\textbf{ACC}} & \multicolumn{1}{c|}{84.69}  & \multicolumn{1}{c|}{86.00}  & \multicolumn{1}{c|}{86.21}  & \multicolumn{1}{c|}{92.86}  & \multicolumn{1}{c|}{93.14}  & \multicolumn{1}{c|}{94.68}  \\ \hline
\multicolumn{7}{c}{Low BPP ($\sim$0.04 BPP)}                                                                                                                                                                           \\ \hline
\multicolumn{1}{|c|}{}             & \multicolumn{3}{c|}{\textbf{Fire DS}}                                                   & \multicolumn{3}{c|}{\textbf{Crack DS}}                                                  \\ \hline
\multicolumn{1}{|c|}{\textbf{BPP}} & \multicolumn{1}{c|}{0.0285}  & \multicolumn{1}{c|}{0.0437}  & \multicolumn{1}{c|}{0.0535}  & \multicolumn{1}{c|}{0.0429} & \multicolumn{1}{c|}{0.0441} & \multicolumn{1}{c|}{0.0459} \\ \hline
\multicolumn{1}{|c|}{\textbf{ACC}} & \multicolumn{1}{c|}{93.46}  & \multicolumn{1}{c|}{94.18}  & \multicolumn{1}{c|}{95.25}  & \multicolumn{1}{c|}{97.91}  & \multicolumn{1}{c|}{97.92}  & \multicolumn{1}{c|}{97.98}  \\ \hline
\end{tabular}
\end{table}

\subsubsection{Infrastructure Analysis}
We construct two randomly sampled, disjoint sets of 2000 images for both knowledge acquisition and evaluation, respectively. 1000 images from the positive (with crack) class and another 1000 images from the negative (no crack) class are present in each of these sets. For knowledge acquisition parameters~(Algo.~\ref{alg:calibration}), we use block dimension ($bDim$) 64, number of iteration ($iterLimit$) 10, prune window size ($pw$) (4 and 8), grid dimension ($grid$) $ceil((rows+cols)/20)$, triangle standard deviation splitting threshold ($th$) 5, and $cb$ 8.

We compress the sampled 2000 evaluation images using MAGIC with compression parameters~(Algo.~\ref{alg:encode_main}) block dimension ($bDim$) 64, $d$ (1 up to 12 in separate instances), grid dimension ($grid$) $ceil((rows+cols)/20)$ along with the domain-specific point prediction model ($model$) and the color quantization dictionary obtained. To compare with MAGIC, we compress the same images with JPEG 2000 and WebP from QF 1 to 10.

\begin{figure}[]
\centering
\includegraphics[width=\columnwidth]{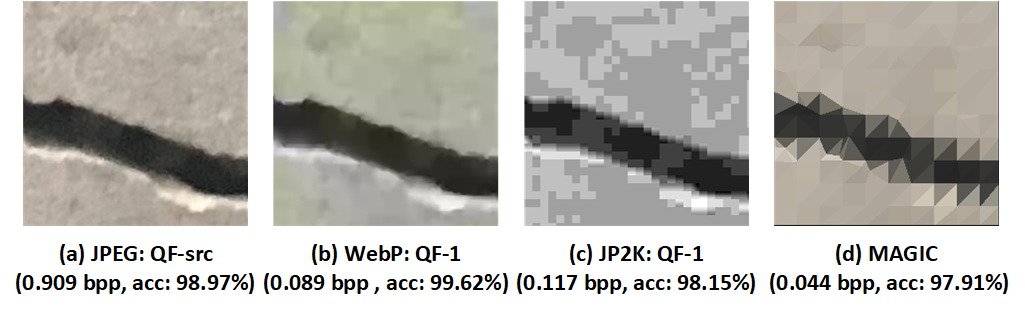}
\caption{Comparison of source, WebP, JPEG 2000, and MAGIC compressed building crack images.}
\label{BuildingCrackDS}

\end{figure}

We obtain a separate dataset for each JPEG 2000, WebP, and MAGIC setting. Fig.~\ref{BuildingCrackDS} shows sample images from the compressed datasets. For each dataset, we extract the features from the second fully connected (fc2) layer of pretrained VGG-16 \cite{ref:vgg} to train and test a support vector machine for the classification task using 30-fold cross-validation (20/80 test/train splits). From Fig.~\ref{CrackAcc}, MAGIC was able to compress beyond JPEG 2000 QF=1 while maintaining almost similar classification accuracy. The MAGIC images in the dataset compressed with $d = 12$ and $pw = 8$ are on average 22.09x smaller (1.06\% accuracy loss) than source dataset (ACC=98.97\%, BPP=0.9479), 2.51x smaller (0.24\% accuracy loss) than JPEG 2000 QF=1 (ACC=98.15\%, BPP=0.1080), and 1.98x smaller  (1.69\% accuracy loss) than WebP QF=1 (ACC=99.60\%, BPP=0.0851).

\subsubsection{Forest Surveillance}
From the forest fire dataset \cite{ref:fireDS}, we extract 643 images of which 227 have fire and 416 have no fire. We ignore the images which are not relevant to forests. We use 20 images from the dataset (10 from each class) to perform the knowledge acquisition procedure. As knowledge acquisition parameters~(Algo.~\ref{alg:calibration}) we use block dimension ($bDim$) 64, number of iteration ($iterLimit$) 10, prune window size ($pw$) (5 and 8), grid dimension ($grid$) $ceil((rows+cols)/20)$, triangle standard deviation splitting threshold ($th$) 5 and $cb$ 8. The domain-specific point prediction model is trained in the same manner as for the infrastructure analysis task. 
 We compress the remaining 623 images (excluding the knowledge acquisition learning set) using MAGIC with compression parameters~(Algo.~\ref{alg:encode_main}) block dimension ($bDim$) 64, $d$ (1 through 12), grid dimension ($grid$) $ceil((rows+cols)/20)$ along with the domain-specific point prediction model ($model$) and the color quantization dictionary obtained from the knowledge acquisition stage. 

Again, we obtain a separate dataset for each JPEG 2000 (QF 1 to 10), WebP (QF 1 to 10), and MAGIC settings. Fig.~\ref{fireDS} shows sample images from the fire dataset for JPEG 2000, WebP, and MAGIC. We extract the features for each dataset similar to the building crack dataset and carry out classification using a support vector machine with 30-fold cross-validation (20/80 test/train splits).
As seen in Fig.~\ref{fireAcc}, we observe the same trend from the previous dataset. The MAGIC images compressed with $d = 8$ and $pw = 8$ are on average 42.65x smaller (2.99\% accuracy loss) than source dataset (ACC=97.17\%, BPP=1.864), 2.32x smaller (1.20\% accuracy loss) than JPEG 2000 QF=1 (ACC=95.38\%, BPP=0.1014), and 5.85x smaller (3.18\% accuracy loss) than WebP QF=1 (ACC=97.36\%, BPP=0.2559).

\subsection{Evaluation beyond Lowest Quality Factor}
WebP and JPEG 2000 are unable to compress beyond QF=1 without some level of pre-processing. On the other hand, MAGIC naturally can achieve a very large compression range. In Table~\ref{PushMore}, we evaluate MAGIC at extreme levels of compression using smaller $cb$ bit sizes. We can compress up to $\sim$167x more than source at $\sim$13\% accuracy loss for the fire dataset and $\sim$69x more than source at $\sim$6\% accuracy loss for the building crack dataset. Depending on the application requirements, MAGIC can gracefully trade-off accuracy for lower BPP using the parameters exposed to the user. This extreme level of compression is possible due to MAGIC's ability to leverage domain knowledge.

\subsection{MAGIC Time \& Energy Analysis}

\begin{table*}[]
\centering
\caption{C/T Cutoff of different MAGIC settings for Building Crack detection dataset \cite{ref:crackDS}.}
\label{buld_fullPower_res}
\renewcommand{\arraystretch}{1.5}
\scriptsize\addtolength{\tabcolsep}{5pt}
\begin{tabular}{|c|c|c|c|c|c|c|c|}
\hline
                                                                               & \multicolumn{4}{c|}{\textbf{MAGIC}}                                                                                                                                                                                                            & \textbf{Source} & \textbf{JP2K} & \textbf{WebP} \\ \hline
\textbf{\begin{tabular}[c]{@{}c@{}}Data\\ Compression Parameters\end{tabular}} & \begin{tabular}[c]{@{}c@{}}pw=8\\ cb=2\\ d=1\end{tabular} & \begin{tabular}[c]{@{}c@{}}pw=8\\ cb=8\\ d=2\end{tabular} & \begin{tabular}[c]{@{}c@{}}pw=4\\ cb=8\\ d=10\end{tabular} & \begin{tabular}[c]{@{}c@{}}pw=4\\ cb=8\\ d=2\end{tabular} & -               & QF-1          & QF-1          \\ \hline
\textbf{Avg BPP}                                                               & 0.012                                                     & 0.038                                                     & 0.039                                                      & 0.046                                                     & 0.947           & 0.108         & 0.085         \\ \hline
\textbf{Avg Size (Byte)}                                                       & 81.83                                                     & 248.5                                                     & 257.2                                                      & 301.0                                                     & 8351            & 696.0         & 548.89        \\ \hline
\textbf{Encode + Setup Time (sec)}                                             & 1.706                                                     & 1.663                                                     & 1.717                                                      & 1.719                                                     & 0               & 0.0083        & 0.0199        \\ \hline
\textbf{Encode Time (sec)}                                                     & 0.090                                                     & 0.098                                                     & 0.101                                                      & 0.103                                                     & 0               & 0.0083        & 0.0199        \\ \hline
\textbf{Decode Time (sec)}                                                     & 0.018                                                     & 0.016                                                    & 0.017                                                     & 0.018                                                     & 0               & 0.0106        & 0.0219        \\ \hline
\textbf{Detection Accuracy  (\%)}                                              & 91.89                                                     & 97.52                                                     & 97.18                                                      & 97.53                                                     & 98.97           & 98.15         & 99.6          \\ \hline
\textbf{C/T Cutoff Source ($1\times10^6$ CC / Byte)}                                        & 0.040                                                     & 0.044                                                     & 0.046                                                      & 0.047                                                     & -               & -             & -             \\ \hline
\textbf{C/T Cutoff JP2K-QF1 ($1\times10^6$ CC / Byte)}                                      & 0.492                                                     & 0.741                                                     & 0.781                                                      & 0.887                                                      & -               & -             & -             \\ \hline
\textbf{C/T Cutoff WebP-QF1 ($1\times10^6$ CC / Byte)}                                      & 0.555                                                     & 0.961                                                     & 1.028                                                      & 1.240                                                     & -               & -             & -             \\ \hline
\end{tabular}
\end{table*}

MAGIC, as per its current implementation, takes longer time to compress images as compared to JPEG 2000 and WebP. However, as shown above, MAGIC can achieve a higher compression rate while still performing well when it comes to coarse-grained machine vision classification tasks. To explore the potential energy savings of MAGIC compression, we introduce a threshold, C/T Cutoff (inspired by Sadler et al.~\cite{sadler2006data}), for determining the sufficient computation and transmission energy consumption ratio beyond which MAGIC will be beneficial for overall energy consumption in a given resource-constrained computing system. The C/T Cutoff for MAGIC compression (for a specific set of parameters) can be computed using the Equation~\ref{eq1} where $E_1$ is the average MAGIC encoding time, $E_2$ is the average encoding time of the competitor method (JPEG 2000, WebP), $I_1$ is the average image size of MAGIC, $I_2$ is the average image size of the competitor method (JPEG 2000, WebP) and $f$ is the CPU clock frequency. The setup time during encoding is due to loading the libraries and initializing the Python environment. In an amortized analysis for a batch operation, the setup time can be considered negligible. 
For MAGIC compression (for a specific set of parameters) to save energy when compared to other compression standards, the operating device must have a C/T value greater than MAGIC's C/T Cutoff. In Tables~\ref{buld_fullPower_res}, \ref{fire_fullPower_res}, we see the C/T Cutoffs for different MAGIC compression settings for building crack detection and forest fire detection datasets, respectively. We use $f$ = 3.7 GHz for computing the C/T cutoff values. Any device with C/T value greater than the cutoff will benefit (in terms of operational power consumption) from using MAGIC with respect to the method being compared against (JPEG 2000, WebP). For example in Table~\ref{fire_fullPower_res}, with MAGIC (pw=8, cb=2, d=1) the JPEG 2000 (JP2K) C/T cutoff is 0.497 which means the energy for 1 byte transmission must be greater than the execution energy of 0.497 million clock cycles (CC) in a system for MAGIC to have higher energy savings than JPEG 2000. 

\begin{equation}
\label{eq1}
C/T\_Cutoff = \lVert \frac{E_1 - E_2} {I_1 - I_2}\rVert  * f
\end{equation}

\section{Discussion}\label{sec:discussion}
In this section, we investigate the properties of the current embodiment of MAGIC. We note, the MAGIC framework can be improved across many dimensions. We make preliminary studies of these possibilities and explore future extensions of MAGIC in this section.

\subsection{Variation in Compression Ability}

\begin{figure}[]
\centering
\includegraphics[width=0.92\columnwidth]{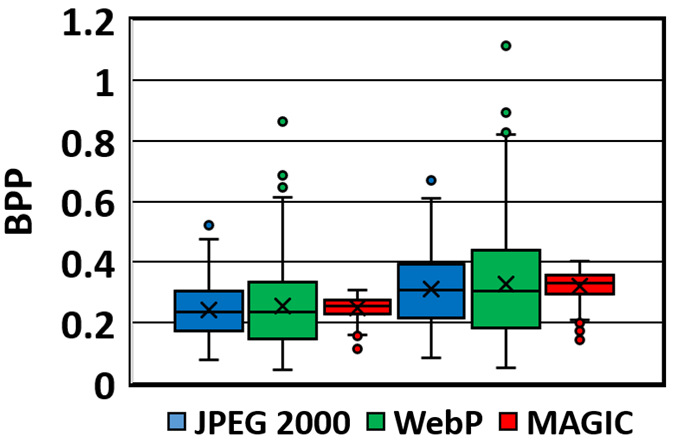}
\caption{Comparison of BPP variance for MAGIC, JPEG~2000, and WebP for building crack dataset.}
\label{crackbppbox}

\end{figure}

\begin{figure}[]
\centering
\includegraphics[width=0.92\columnwidth]{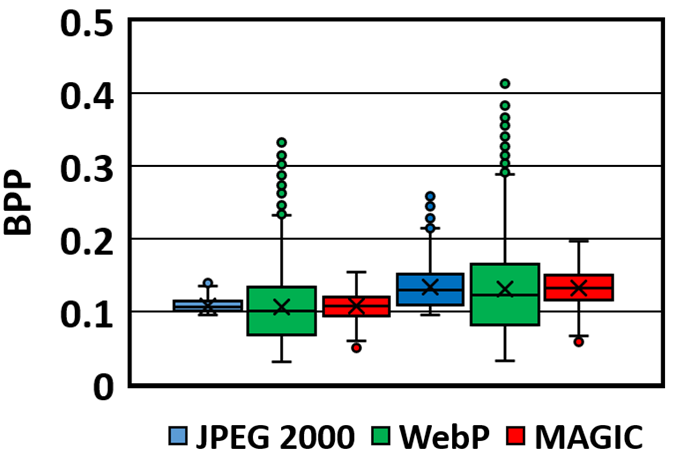}
\caption{Comparison of BPP variance for MAGIC, JPEG~2000 and WebP for fire dataset.}
\label{firebppbox}

\end{figure}

The image compression technique being used must generate images of less size variability for maintaining consistent overall system performance. We compress the images using JPEG 2000, WebP and MAGIC to generate box plots showing the variation of BPP for the sampled distributions in Fig.~\ref{firebppbox} and Fig.~\ref{crackbppbox}. We observe that MAGIC provides low variation in BPP as compared to JPEG 2000 and WebP images. Due to different parameters in the knowledge acquisition and encoding phase, specifically $pw$, MAGIC has fine control over the compressed image size. Hence, MAGIC can provide steady performance even in biased scenarios, where other techniques may not give good compression.

 \begin{table*}[]
\centering
\caption{C/T Cutoff of different MAGIC settings for Forest Fire detection dataset \cite{ref:fireDS}.}
\label{fire_fullPower_res}
\renewcommand{\arraystretch}{1.5}
\scriptsize\addtolength{\tabcolsep}{5pt}
\begin{tabular}{|c|c|c|c|c|c|c|c|}
\hline
                                                                                  & \multicolumn{4}{c|}{\textbf{MAGIC}}                                                                                                                                                                                                             & \textbf{Source} & \textbf{JP2K} & \textbf{WebP} \\ \hline
\textbf{\begin{tabular}[c]{@{}c@{}}Dataset\\ Compression Parameters\end{tabular}} & \begin{tabular}[c]{@{}c@{}}pw=8\\ cb=8\\ d=12\end{tabular} & \begin{tabular}[c]{@{}c@{}}pw=8\\ cb=8\\  d=4\end{tabular} & \begin{tabular}[c]{@{}c@{}}pw=8\\ cb=8\\ d=3\end{tabular} & \begin{tabular}[c]{@{}c@{}}pw=8\\ cb=8\\ d=1\end{tabular} & -               & QF-1          & QF-1          \\ \hline
\textbf{Avg BPP}                                                                  & 0.019                                                      & 0.029                                                      & 0.034                                                     & 0.074                                                     & 1.864           & 0.101         & 0.013         \\ \hline
\textbf{Avg Size (Byte)}                                                          & 431.3                                                      & 658.0                                                      & 771.2                                                     & 1675                                                      & 44102           & 2282          & 5756          \\ \hline
\textbf{Encode + Setup Time (sec)}                                                & 1.830                                                      & 1.841                                                      & 1.890                                                     & 1.885                                                     & 0               & 0.0097        & 0.036         \\ \hline
\textbf{Encode Time (sec)}                                                        & 0.261                                                      & 0.272                                                      & 0.281                                                     & 0.315                                                    & 0               & 0.0097        & 0.036         \\ \hline
\textbf{Decode Time (sec)}                                                        & 0.032                                                      & 0.034                                                      & 0.035                                                     & 0.039                                                    & 0               & 0.0209        & 0.084         \\ \hline
\textbf{Detection Accuracy (\%)}                                                  & 90.56                                                      & 92.72                                                      & 93.2                                                      & 94.92                                                     & 97.17           & 95.38         & 97.36         \\ \hline
\textbf{C/T Cutoff Source ($1\times10^6$ CC / Byte)}                                           & 0.022                                                      & 0.023                                                      & 0.023                                                     & 0.027                                                     & -               & -             & -             \\ \hline
\textbf{C/T Cutoff JP2K-QF1($1\times10^6$ CC / Byte)}                                          & 0.502                                                      & 0.597                                                      & 0.664                                                     & 1.860                                                     & -               & -             & -             \\ \hline
\textbf{C/T Cutoff WebP-QF1($1\times10^6$ CC / Byte)}                                          & 0.156                                                      & 0.171                                                      & 0.181                                                     & 0.252                                                     & -               & -             & -             \\ \hline
\end{tabular}
\end{table*}

\subsection{Improving Prediction Accuracy }
Post-processing the MAGIC images or using a more powerful pattern prediction model can improve the prediction accuracy by about 1-2\%. Images compressed using MAGIC consist of triangulation artifacts. One way to remove the artifacts is to recursively subdivide the triangles and compute the approximate color of each sub triangle based on the colors of the triangle and its neighbours. Using this technique, we were able to increase the classification accuracy. However, there will be extra computation due to post-processing in the decoder end. If the decoder system resides in the cloud, then this step can be considered to squeeze out extra performance.

As explained earlier, we use entropy features for training and using our neural network models, but we have noticed that VGG-16 fc2 features perform slightly better. Using a VGG inspired large convolution neural network for carrying out the domain-specific point prediction task also improves the performance slightly. However, we intentionally use simple entropy features and a small neural network to boost speed, and help reduce energy consumption and space requirements. In an application where time, space, and energy are not constrained, we can opt for more complex feature extraction methods and larger neural network architectures for domain-specific point prediction.

\subsection{Time Complexity and Performance Improvements}

Time complexity analysis of encoder (Algo.~\ref{alg:encode_main}) and decoder (Algo.~\ref{alg:decode_main}) algorithms simplify to  $\mathcal{O}(N+M\log M+TR)$. The major contributors in encoding are  $\mathcal{O}(N)$ for tiling (line 2),  $\mathcal{O}(M\log M)$ for DT (line 14), and  $\mathcal{O}(TR)$ for triangle color calculation (line 17, the pixels associated with a triangle are determined by searching in a rectangle circumscribing the triangle), where $N$ is the number of pixels in the image, $M$ is the number of points sprayed, $T$ is the number of triangles, and $R$ is the dimension of the bounding rectangle of the biggest triangle. 
For decoding, the contributors are  $\mathcal{O}(N)$ for predicted point absolute position computation (lines 6-8),  $\mathcal{O}(M\log M)$ for DT (line 12), and  $\mathcal{O}(TR)$ for triangle color assignment/drawing (line 17). In both algorithms, we expect the  $\mathcal{O}(M\log M)$ DT step to consume the most time. 

Time complexity analysis of the knowledge acquisition algorithm (Algo.~\ref{alg:calibration}) simplifies to $\mathcal{O}(KN\log N + KIM\log M + KITR + SVC + PQ)$. The major contributors are  $\mathcal{O}(KN\log N)$ for canny edge detection for all $K$ images (line 7), $\mathcal{O}(KIM\log M + KITR)$ for the split operation across all $K$ images (line 11), $\mathcal{O}(SVC)$ for color dictionary computation using k-means algorithm (line 29)  and, $\mathcal{O}(PQ)$ for training the point prediction model (line 30). $N$, $M$, $T$, $R$ hold the same meaning as before and additionally $K$ is the number of images in the $imgList$, $I$ is the $iterLimit$, $S$ is the iteration limit for k-means algorithm, $V$ is the number of points in the $colorFreq$ map, $C$ is the number of centroids specified for k-means, $P$ is the number of training samples in $trainX$ and $trainY$, $Q$ is the number of training epochs for the point prediction model.

The runtime performance of both decoder and encoder can be improved through parallelization, hardware implementation and code tweaking. Many of the block operations such as block feature extraction and point spray pattern prediction can be easily parallelized. Hardware implementation can provide the most speed up and may help reduce energy consumption as well. In future works, we will focus on improving the time and energy performance of MAGIC using different means.
  
\subsection{Manual Region-of-Interest Guided Compression}

\begin{figure}[]
\centering
\includegraphics[width=0.8\columnwidth]{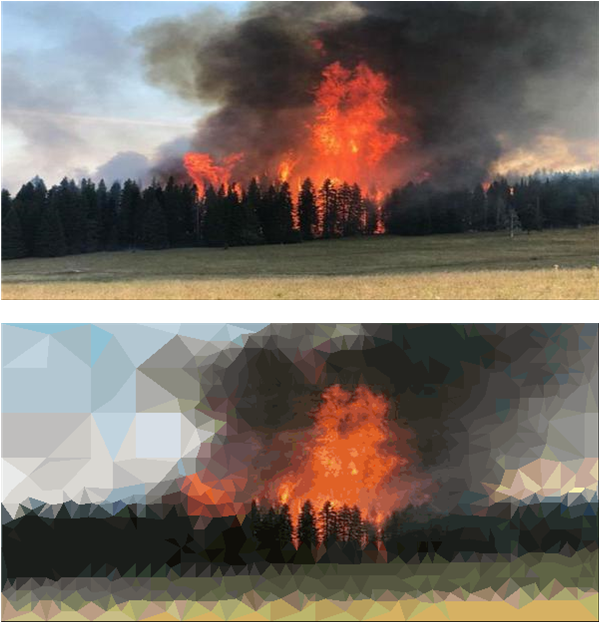}
\caption{Region-of-Interest Guided MAGIC Compression on a sample Fire Detection Dataset image \cite{ref:fireDS}.}
\label{ROI_Example_MAGIC}

\end{figure}

As previously described, ROI bias can be automatically captured by training the pattern prediction model with supervised images. Beyond learning the region-of-interest bias, MAGIC offers manual ROI based compression. With this feature, users can specify additional regions of an image that can be retained at a higher quality. In Fig.~\ref{ROI_Example_MAGIC}, we see an example ROI-guided image compression where the fire region is designated manually as a region-of-interest. Note that the image region with the fire maintains much higher information than the remaining regions.

\subsection{Extension of MAGIC to video compression}
A video can be thought of as a collection of images. To this end, MAGIC can be extended to process videos as well. Depending on the sampling rate of the image sensor, we noticed that, adjacent video frames have very little content difference. Taking this into consideration, we can save more in terms of space, computation and transmission. The two main components of a MAGIC encoded image are $labelsArr$ and $colorDict$. We can represent frame[$N$] by reusing the $colorDict$ and $labelArr$ of frame[$N-1$]. In Eqn.~\ref{eq2}, $OP$ is the set of obsolete point spray patterns which are no longer present in the new frame and  $NP$ is the set of new point spray patterns which are introduced in the new frame. Similarly, as shown in Eqn.~\ref{eq3}, the $colorDict$[$N-1$] can be modified by removing the obsolete triangle colors and introducing the colors of the new triangles in frame $N$. Future works will investigate and formalize the MAGIC flow applied to video.
\begin{equation}
\label{eq2}
labelsArr[N] = labelsArr[N-1] - OP + NP 
\end{equation}
\begin{equation}
\label{eq3}
colorDict[N] = colorDict[N-1] - OC + NC
\end{equation}

\section{Conclusion}
\label{sec:conclusion}

The increasing use of intelligent edge devices in diverse IoT applications, specifically for a multitude of computer vision tasks, calls for innovation in an image compression technique that meets the unique requirements of these applications. They primarily require high compression ratio while maintaining machine vision accuracy at an acceptable level. The MAGIC framework we have presented in this paper addresses this need. We have shown that effective use of domain knowledge learned with ML can provide high compression in resource-constrained edge applications while keeping appropriate features for machine vision. The proposed framework is flexible for application in diverse domains and scalable to large image sizes. Our experiments for coarse-grained ML tasks using two datasets highlight the effectiveness of MAGIC. We achieve up to 42.65x higher compression than the source (beyond JPEG 2000 and WebP) while achieving similar accuracy. We compute the transmission computation energy cutoffs to demonstrate at what level MAGIC compressed images can be more energy efficient than standard techniques. Further, we show low compression variance compared to standard image compression techniques. With the use of a common pattern dictionary, the proposed ML-based compression procedure can be easily extended for recognizing coarse-grain patterns in edge devices. Moreover, it can potentially be extended to video compression where domain knowledge is expected to play an even stronger role. Future work will investigate these extensions as well as further improvement in the performance of MAGIC in a variety of edge applications.





\bibliographystyle{IEEEtran}
\bibliography{IEEEabrv,bib}

\end{document}